\documentclass[11pt,a4paper]{article}
\usepackage[hyperref]{emnlp2018}
\usepackage{times}
\usepackage{latexsym}

\usepackage{url}

\usepackage[greek, english]{babel}
\usepackage[utf8]{inputenc} % allow utf-8 input
\usepackage{booktabs}       % professional-quality tables
\usepackage{amsfonts}       % blackboard math symbols
\usepackage{dsfont} % Double stroke fonts

\usepackage{xspace}

\usepackage{bbm}          % bold math symbols

\usepackage{amssymb,amsmath,amsthm, mathtools}
\usepackage{amscd}

\usepackage{caption, subcaption}%, subcaption}% http://ctan.org/pkg/{caption,subcaption}

\usepackage{siunitx} % scientific

\usepackage{algorithm}
\usepackage{algorithmic}

\usepackage[colorinlistoftodos]{todonotes}
\newtheorem{theorem}{Theorem}[section]

\newtheorem{lemma*}{Lemma}

\makeatletter
\renewcommand*\env@matrix[1][*\c@MaxMatrixCols c]{%
  \hskip -\arraycolsep
  \let\@ifnextchar\new@ifnextchar
  \array{#1}}
\makeatother

\def\R{\mathbb{R}}
\def\K{\mathbb{K}}

\def\C{\mathbf{C}}

\def\I{\mathbf{I}}
\def\K{\mathbf{K}}
\def\L{\mathbf{L}}
\def\P{\mathbf{P}}
\def\U{\mathbf{U}}
\def\V{\mathbf{V}}
\def\Y{\mathbf{Y}}
\def\X{\mathbf{X}}

\def\x{\mathbf{x}}

\def\y{\mathbf{y}}

\def\p{\mathbf{p}}
\def\q{\mathbf{q}}

\def\b{\mathbf{b}}

\def\a{\mathbf{a}}

\def\0{\mathbf{0}}
\def\one{\mathbbm{1}}
\def\cX{\mathcal{X}}

\def\cF{\mathcal{F}}

\def\ones{\mathds{1}}

\def\st{\medspace|\medspace}

\DeclarePairedDelimiterX{\inner}[2]{\langle}{\rangle}{#1, #2}

\DeclareDocumentCommand{\tr}{s m}{% \Pr[*]{..}
  \operatorname{tr}%
  \IfBooleanTF{#1}% Condition on *
    {#2}% Print only the argument in starred * version
    {\left(#2\right)}% Print bracketed argument [ ] in unstarred version
}%
\DeclareDocumentCommand{\diag}{s m}{% \Pr[*]{..}
  \operatorname{diag}%
  \IfBooleanTF{#1}% Condition on *
    {#2}% Print only the argument in starred * version
    {\left(#2\right)}% Print bracketed argument [ ] in unstarred version
}%

\def\gw{\text{GW}}

\DeclarePairedDelimiterX{\infdivx}[2]{(}{)}{%
  #1\;\delimsize\|\;#2%
}

\newcommand{\suchthat}{\;\ifnum\currentgrouptype=16 \middle\fi|\;}

\newenvironment{itemize*}%
  {\begin{itemize}%
  \vspace{-0.5cm}
    \setlength{\itemsep}{0pt}%
    \setlength{\parskip}{0pt}}%
  {\end{itemize}}
\newenvironment{enumerate*}%
{\begin{enumerate}
    \vspace{-0.5cm}
    \setlength{\itemsep}{0pt}%
    \setlength{\parskip}{0pt}}%
  {\end{enumerate}}

\newcommand{\ours}{\textsc{G-W}\xspace}
\newcommand{\procru}{\textsc{Procrustes}\xspace}

\newcommand{\En}{\textsc{En}\xspace}
\newcommand{\Es}{\textsc{Es}\xspace}
\newcommand{\De}{\textsc{De}\xspace}
\newcommand{\Fi}{\textsc{Fi}\xspace}
\newcommand{\Fr}{\textsc{Fr}\xspace}
\newcommand{\It}{\textsc{It}\xspace}
\newcommand{\Ru}{\textsc{Ru}\xspace}
\newcommand{\Zh}{\textsc{Zh}\xspace}
\newcommand{\Eo}{\textsc{Eo}\xspace}

  {\list{}{\leftmargin=#1\rightmargin=#1}\item[]}%
  {\endlist}

\aclfinalcopy % Uncomment this line for the final submission

\title{Gromov-Wasserstein Alignment of Word Embedding Spaces}

\author{David Alvarez-Melis \\
  CSAIL, MIT \\
  {\tt dalvmel@mit.edu} \\\And
  Tommi S. Jaakkola \\
  CSAIL, MIT \\
  {\tt tommi@mit.edu} \\}
\date{}

\begin{document}
\selectlanguage{english}
\maketitle

%!TEX root = main.tex

\begin{abstract}

Cross-lingual or cross-domain correspondences play key roles in tasks ranging from machine translation to transfer learning. Recently, purely unsupervised methods operating on monolingual embeddings have become effective alignment tools. Current state-of-the-art methods, however, involve multiple steps, including heuristic post-hoc refinement strategies. In this paper, we cast the correspondence problem directly as an optimal transport (OT) problem, building on the idea that word embeddings arise from metric recovery algorithms. Indeed, we exploit the \emph{Gromov-Wasserstein} distance that measures how similarities between pairs of words relate across languages. We show that our OT objective can be estimated efficiently, requires little or no tuning, and results in performance comparable with the state-of-the-art in various unsupervised word translation tasks. 

\end{abstract}

\section{Introduction}

Many key linguistic tasks, within and across languages or domains, including machine translation, rely on learning cross-lingual correspondences between words or other semantic units. While the associated alignment problem could be solved with access to large amounts of parallel data, broader applicability relies on the ability to do so with largely mono-lingual data, from Part-of-Speech (POS) tagging \citep{zhang2016ten}, dependency parsing \cite{guo2015cross}, to machine translation \cite{lample2018unsupervised}. The key subtask of bilingual lexical induction, for example, while long standing as a problem \cite{fung1995compiling, rapp1995identifying, rapp1999automatic}, has been actively pursued recently \citep{artetxe2016learning, zhang2017adversarial, conneau2018word}. 

Current methods for learning cross-domain correspondences at the word level rely on distributed representations of words, building on the observation that mono-lingual word embeddings exhibit similar geometric properties across languages \citet{Mikolov2013Exploiting}. While most early work assumed some, albeit minimal, amount of parallel data \citep{Mikolov2013Exploiting, dinu2014improving, zhang2016ten}, recently fully-unsupervised methods have been shown to perform on par with their supervised counterparts \cite{conneau2018word, Artetxe2018Robust}. While successful, the mappings arise from multiple steps of processing, requiring either careful initial guesses or post-mapping refinements, including mitigating the effect of frequent words on neighborhoods. The associated adversarial training schemes can also be challenging to tune properly \cite{Artetxe2018Robust}.

In this paper, we propose a direct optimization approach to solving correspondences based on recent generalizations of optimal transport (OT). OT is a general mathematical toolbox used to evaluate correspondence-based distances and establish mappings between probability distributions, including discrete distributions such as point-sets. However, the nature of mono-lingual word embeddings renders the classic formulation of OT inapplicable to our setting. Indeed, word embeddings are estimated primarily in a relational manner to the extent that the algorithms are naturally interpreted as metric recovery methods \cite{Hashimoto2016Word}. In such settings, previous work has sought to bypass this lack of \emph{registration} by jointly optimizing over a matching and an orthogonal mapping \cite{Rangarajan1997Softassign, zhang2017earth}. Due to the focus on distances rather than points, we instead adopt a relational OT formulation based on the Gromov-Wasserstein distance that measures how distances between pairs of words are mapped across languages. We show that the resulting mapping admits an efficient solution and requires little or no tuning. 

In summary, we make the following contributions:
\begin{itemize}
	\item We propose the use of the Gromov-Wasserstein distance to learn correspondences between word embedding spaces in a fully-unsupervised manner, leading to a  theoretically-motivated optimization problem that can be solved efficiently, robustly, in a single step, and requires no post-processing or heuristic adjustments.
	\item To scale up to large vocabularies we realize an extended mapping to words not part of the original optimization problem.	
	\item We show that the proposed approach performs on par with state-of-the-art neural network based methods on benchmark word translation tasks, while requiring a fraction of the computational cost and/or hyper-parameter tuning.
\end{itemize}

%%% Local Variables:
%%% mode: latex
%%% TeX-master: "main"
%%% End:
			% Abstract + Intro
%!TEX root = main.tex

\section{Problem Formulation}

In the unsupervised bilingual lexical induction problem we consider two languages with vocabularies $V_x$ and $V_y$, represented by word embeddings $X=\{\x^{(i)}\}_{i=1}^n$ and $Y=\{\y^{(j)}\}_{j=1}^m$, respectively, where $\x^{(i)}\in \mathcal{X} \subset \R^{d_x}$ corresponds to $w_i^x \in V_x$ and $\y^{(j)}\in \mathcal{Y} \subset\R^{d_y}$ to $w_j^y \in V_y$. For simplicity, we let $m=n$ and $d_x = d_y$, although our methods carry over to the general case with little or no modifications. Our goal is to learn an alignment between these two sets of words without any parallel data, i.e., we learn to relate $\x^{(i)} \leftrightarrow \y^{(j)}$ with the implication that $w_i^x$ translates to $w_j^y$. 

As background, we begin by discussing the problem of learning an explicit map between embeddings in the supervised scenario. The associated training procedure will later be used for extending unsupervised alignments (Section~\ref{sub:scaling_up}).

\subsection{Supervised Maps: Procrustes} % (fold)
\label{sub:subsection_name}
In the supervised setting, we learn a map $T:\mathcal{X}\rightarrow \mathcal{Y}$ such that $T(\x^{(i)}) \approx \y^{(j)}$ whenever $w^y_j$ is a translation of $w^x_i$. Let $\X$ and $\Y$ be the matrices whose columns are vectors $\x^{(i)}$ and $\y^{(j)}$, respectively. Then we can find $T$ by solving
\begin{equation}
	\min_{T \in \mathcal{F}} \| \X - T(\Y) \|_F^2
\end{equation}
where $\| \cdot\|_F$ is the Frobenius norm $\|A\|_F = \sqrt{\sum_{i,j}|a_{ij}|^2}$.
Naturally, both the difficulty of finding $T$ and the quality of the resulting alignment depend on the choice of space $\cF$. A classic approach constrains $T$ to be orthonormal matrices, i.e., rotations and reflections, resulting in the orthogonal Procrustes problem
\begin{equation}\label{eq:procrustes}
	\min_{\P \in O(n)} \| \X - \P\Y \|_F^2
\end{equation}
where $O(n) = \{ \P \in \R^{n\times n} \suchthat \P^\top\P = \I\}$. One key advantage of this formulation is that it has a closed-form solution in terms of a singular value decomposition (SVD), whereas for most other choices of constraint set $\mathcal{F}$ it does not. Given an SVD decomposition $\U\Sigma\V^\top$ of $\X\Y^\top$, the solution to problem \eqref{eq:procrustes} is
$
	\P^* =  \U\V^\top
$ \citep{schonemann1966generalized}.
Besides obvious computational advantage, constraining the mapping between spaces to be orthonormal is justified in the context of word embedding alignment because orthogonal maps preserve angles (and thus distances), which is often the only information used by downstream tasks (e.g., for nearest neighbor search) that rely on word embeddings.  \citep{smith2017offline} further show that orthogonality is required for self-consistency of linear transformations between vector spaces.

Clearly, the Procrustes approach only solves the supervised version of the problem as it requires a known correspondence between the columns of $\X$ and $\Y$. Steps beyond this constraint include using small amounts of parallel data \cite{zhang2016ten} or an unsupervised technique as the initial step to generate pseudo-parallel data \citep{conneau2018word} before solving for $\P$. 
\subsection{Unsupervised Maps: Optimal Transport}\label{sub:ot} % (fold)
\label{sub:optimal_transport}
Optimal transport formalizes the problem of finding a minimum cost mapping between two point sets, viewed as discrete distributions. Specifically, we assume two empirical distributions over embeddings, e.g., 
\begin{equation}
	\mu = \sum_{i=1}^n \p_i\delta_{\x^{(i)}}, \quad \nu = \sum_{j=1}^m \q_j\delta_{\y^{(i)}}
\end{equation}
where $\p$ and $\q$ are vectors of probability weights associated with each point set. In our case, we usually consider uniform weights, e.g., $\p_i=1/n$ and $\q_j=1/m$, although if additional information were provided (such as in the form of word frequencies), those could be naturally incorporated via $\p$ and $\q$ (see discussion at the end of Section 3). We find a \emph{transportation map} $T$ realizing
\begin{equation}\label{monge_ot}
	\inf_{T} \left\{ \int_{\cX}c(\x,T(\x))d\mu(\x) \st T_{\#}\mu = \nu \right\},
\end{equation}
where the cost $c(\x,T(\x))$ is typically just $\|\x-T(\x)\|$ and $T_{\#}\mu = \nu$ implies that the source points must exactly map to the targets. However, such a map need not exist in general and we instead follow a relaxed Kantorovich's formulation. In this case, the set of transportation plans is a polytope:
\begin{equation*}%\label{transport_poly}
	\Pi(\p,\q) = \{ \Gamma \in \mathbb{R}^{n \times m}_+ \st \Gamma \ones_n = \p, \medspace \Gamma^\top\ones_n = \q \}.
\end{equation*}
The cost function is given as a matrix $\C \in \R^{n \times m}$, e.g., $C_{ij} = \|\x^{(i)}-\y^{(j)}\|$. The total cost incurred by $\Gamma$ is $\langle \Gamma, C \rangle := \sum_{ij} \Gamma_{ij} C_{ij}$. Thus, the discrete optimal transport (DOT) problem consists of finding a plan $\Gamma$ that solves
\begin{equation}\label{eq:original_OT}
	\min_{\Gamma \in \Pi(\p,\q)} \langle \Gamma, \C \rangle .
\end{equation}
Problem \eqref{eq:original_OT} is a linear program, and thus can be solved exactly in $O(n^3 \log n)$ with interior point methods. However, regularizing the objective leads to more efficient optimization and often better empirical results. The most common such regularization, popularized by \citet{cuturi2013sinkhorn}, involves adding an entropy penalization:
\begin{equation}
\min_{\Gamma \in \Pi(\p,\q) }\; \langle \Gamma, \C \rangle - \lambda H(\Gamma).
\end{equation}
The solution of this strictly convex optimization problem has the form $\Gamma^* = \diag{\a}\K\diag{\b}$, with $\K = e^{-\frac{\C}{\lambda}}$ (element-wise), and can be obtained efficiently via the Sinkhorn-Knopp algorithm, a matrix-scaling procedure which iteratively computes:
\begin{equation}\label{eq:sikhorn_iters}
	\a \gets \p \oslash \K\b \quad \text{and} \quad \b \gets \q \oslash \K^\top \a,
\end{equation}
where $\oslash$ denotes entry-wise division. The derivation of these updates is immediate from the form of $\Gamma^*$ above, combined with the marginal constraints $\Gamma \ones_n = \p, \medspace \Gamma^\top\ones_n = \q$ \citep{Peyre2018Computational}.

Although simple, efficient and theoretically-motivated, a direct application of discrete OT for unsupervised word translation is not appropriate. One reason is that the mono-lingual embeddings are estimated in a relative manner, leaving, e.g., an overall rotation unspecified. Such degrees of freedom can dramatically change the entries of the cost matrix $C_{ij} = \|\x^{(i)}-\y^{(j)}\|$ and the resulting transport map. One possible solution is to simultaneously learn an optimal coupling and an orthogonal transformation \cite{zhang2017earth}. The transport problem is then solved iteratively, using $C_{ij} = \|\x^{(i)}-\P\y^{(j)}\|$, where $\P$ is in turn chosen to minimize the transport cost (via Procrustes). While promising, the resulting iterative approach is sensitive to initialization, perhaps explaining why \citet{zhang2017earth} used an adversarially learned mapping as the initial step. The computational cost can also be prohibitive \citep{Artetxe2018Robust} though could be remedied with additional development. 

We adopt a theoretically well-founded generalization of optimal transport for pairs of points (their distances), thus in line with how the embeddings are estimated in the first place. We explain the approach in detail in the next Section.

\begin{figure*}
	\centering
	\includegraphics[width=\linewidth]{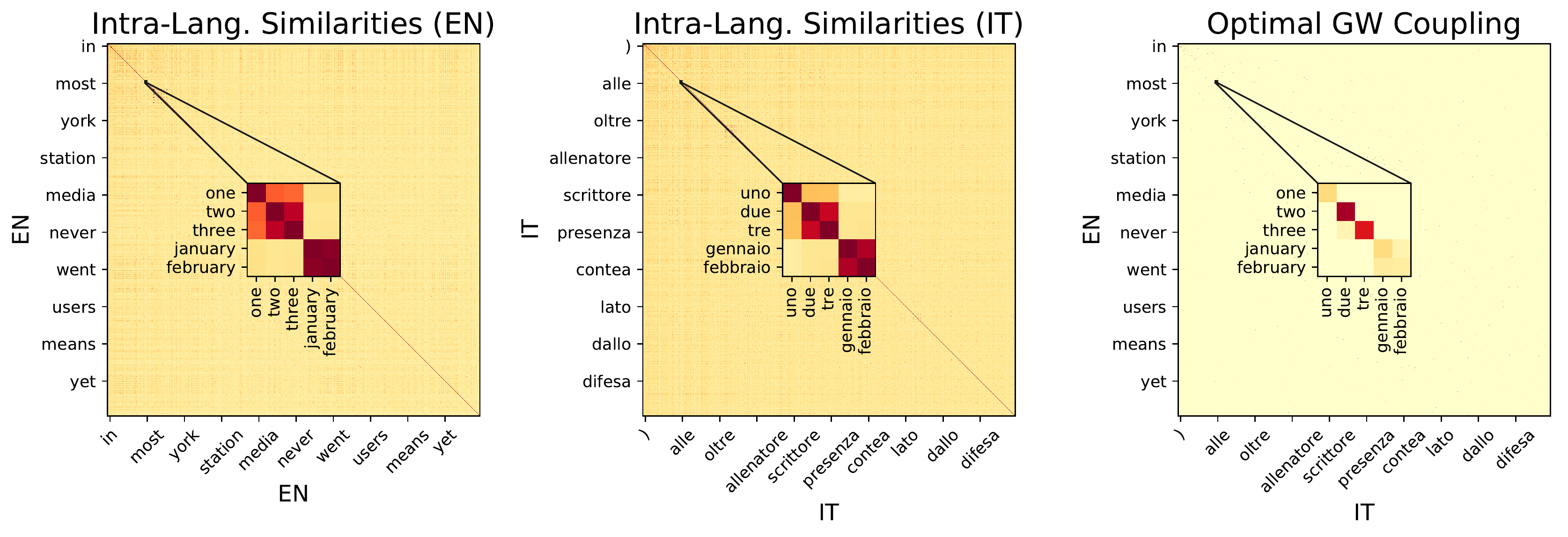}%
	\caption{The Gromov-Wasserstein distance is well suited for the task of cross-lingual alignment because it relies on \emph{relational} rather than \emph{positional} similarities to infer correspondences across domains. Computing it requires two intra-domain similarity (or equivalently cost) matrices (\textbf{left} \& \textbf{center}), and it produces an optimal coupling of source and target points with minimal discrepancy cost (\textbf{right}).}
\end{figure*}

\section{Transporting across unaligned spaces}

In this section we introduce the Gromov-Wasserstein distance, describe an optimization algorithm for it, and discuss how to extend the approach to out-of-sample vectors.

\subsection{The Gromov Wasserstein Distance} % (fold)
\label{sub:the_gromov_wasserstein_distance}

The classic optimal transport requires a distance between vectors \emph{across} the two domains. Such a metric may not be available, for example, when the sample sets to be matched do not belong to the same metric space (e.g., different dimension). The Gromov-Wasserstein distance \citep{memoli2011gromov} generalizes optimal transport by comparing the metric spaces directly instead of samples across the spaces. In other words, this framework operates on distances between pairs of points calculated within each domain and measures how these distances compare to those in the other domain. Thus, it requires a weaker but easy to define notion of \emph{distance between distances}, and operates on pairs of points, turning the problem from a linear to a quadratic one.

Formally, in its discrete version, this framework considers two measure spaces expressed in terms of within-domain similarity matrices $(\C, \p)$ and $(\C', \q)$ and a loss function defined between \emph{similarity pairs}: $L: \R\times\R \rightarrow \R$, where $L(C_{ik}, C_{jl}')$ measures the discrepancy between the distances  $d(\x^{(i)}, \x^{(k)})$ and $d'(\y^{(j)}, \y^{(l)})$. Typical choices for $L$ are $L(a,b) = \frac{1}{2}(a - b)^2$ or $L(a, b) = \text{KL}(a | b)$. In this framework, $L(C_{ik}, C'_{jl})$ can also be understood as the cost of ``matching'' $i$ to $j$ \emph{and} $k$ to $l$. 

All the relevant values of $L(\cdot, \cdot)$ can be put in a 4-th order tensor $\L \in \R^{N_1\times N_1 \times N_2 \times N_2}$, where $\L_{ijkl} = L(C_{ik}, C_{jl}')$. As before, we seek a coupling $\Gamma$ specifying how much mass to transfer between each pair of points from the two spaces. The Gromov-Wasserstein problem is then defined as solving
\begin{equation}\label{eq:gromov_wasserstein}
	\hspace{-0.05cm} \text{GW}(\C, \C', \p, \q) = \hspace{-0.2cm}\min_{\Gamma \in \Pi(\p,\q)} \sum_{i,j,k,l} \L_{ijkl} \Gamma_{ij} \Gamma_{kl} \hspace{-0.1cm}
\end{equation}
Compared to problem \eqref{eq:original_OT}, this version is substantially harder since the objective is now not only non-linear, but non-convex too.\footnote{In fact, the discrete (Monge-type) formulation of the problem is essentially an instance of the well-known (and NP-hard) quadratic assignment problem (QAP).} In addition, it requires operating on a fourth-order tensor, which would be prohibitive in most settings. Surprisingly, this problem can be optimized efficiently with first-order methods, whereby each iteration involves solving a traditional optimal transport problem \citep{peyre2016gromov}. Furthermore, for suitable choices of loss function $L$, \citet{peyre2016gromov} show that instead of the $O(N_1^2N_2^2)$ complexity implied by naive fourth-order tensor product, this computation reduces to $O(N_1^2N_2 + N1N_2^2)$ cost. Their approach consists of solving \eqref{eq:original_OT} by projected gradient descent, which yields iterations that involve projecting onto $\Pi(\p,\q)$ a pseudo-cost matrix of the form
\begin{align}\label{eq:gw_pseudocost}
	\hat{\C}_\Gamma(\C, \C', \Gamma) &= \C_{xy} - h_1(\C)\Gamma h_2(\C')^\top
\end{align}
where
\[ \C_{xy} = f_1(\C)\p\ones_{m}^\top + \ones_n\q^\top f_2(\C')^\top \]
and $f_1, f_2, h_2, h_2$ are functions that depend on the loss $L$. We provide an explicit algorithm for the case $L = L_2$ at the end of this section.

Once we have solved \eqref{eq:gromov_wasserstein}, the optimal transport coupling $\Gamma^*$ provides an explicit (soft) matching between source and target samples, which for the problem of interest can be interpreted as a probabilistic translation: for every pair of words $(w_{src}^{(i)},w_{trg}^{(j)})$, $\Gamma^*_{ij}$ provides a likelihood that these two words are translations of each other. This itself is enough to translate, and we show in the experiments section that $\Gamma^*$ by itself, without any further post-processing, provides high-quality translations. This stands in sharp contrast to mapping-based methods, which rely on nearest-neighbor computation to infer translations, and thus become prone to hub-word effects which have to be mitigated with heuristic post-processing techniques such as Inverted Softmax \citep{smith2017offline} and Cross-Domain Similarity Scaling (\textsc{Csls}) \citep{conneau2018word}. The transportation coupling $\Gamma$, being normalized \emph{by construction}, requires no such artifacts.

The Gromov-Wasserstein problem \eqref{eq:gromov_wasserstein} possesses various desirable theoretical properties, including the fact that for a suitable choice of the loss function it is indeed a distance:
\begin{theorem}[\citealt{memoli2011gromov}]\label{thm:memoli}
	With the choice $L = L_2$, $\text{GW}^{\frac{1}{2}}$ is a distance on the space of metric measure spaces.
\end{theorem}
Solving problem \eqref{eq:gromov_wasserstein} therefore yields a fascinating accompanying notion: the \emph{Gromov-Wasserstein distance between languages}, a measure of semantic discrepancy purely based on the relational characterization of their word embeddings. Owing to Theorem~\ref{thm:memoli}, such values can be interpreted as distances, so that, e.g., the triangle inequality holds among them. In Section~\ref{sub:extra_exp} we compare various languages in terms of their GW-distance.

Finally, we note that whenever word frequency counts are available, those would be used for $\p$ and $\q$. If they are not, but words are sorted according to occurrence (as they often are in popular off-the-shelf embedding formats), one can estimate rank-probabilities such as Zipf power laws, which are known to accurately model multiple languages \cite{Piantadosi2014Zipf}. In order to provide a fair comparison to previous work, throughout our experiments \textbf{we use uniform distributions} so as to avoid providing our method with additional information not available to others.

% subsection the_gromov_wasserstein_distance (end)

\subsection{Scaling Up} % (fold)
\label{sub:scaling_up}

While the pure Gromov-Wasserstein approach leads to high quality solutions, it is best suited to small-to-moderate vocabulary sizes,\footnote{As shown in the experimental section, we are able to run problems of size in the order of $|V_s| \approx 10^5 \approx |V_t|$ on a single machine \textbf{without} relying on GPU computation.} since its optimization becomes prohibitive for very large problems. For such settings, we propose a two-step approach in which we first match a subset of the vocabulary via the optimal coupling, after which we learn an orthogonal mapping through a modified Procrustes problem. 
Formally, suppose we solve problem \eqref{eq:gromov_wasserstein} for a reduced matrices $\X_{1:k}$ and $\Y_{i:k}$ consisting of the first columns $k$ of $\X$ and $\Y$, respectively, and let $\Gamma^*$ be the optimal coupling. We seek an orthogonal matrix that best recovers the barycentric mapping implied by $\Gamma^*$. Namely, we seek to find $\P$ which solves:
\begin{equation}\label{eq:procrustes_gw_problem}
	\min_{\P \in O(n)} \| \X\Gamma^*  - \P\Y\|_2^2
\end{equation}
Just as problem \eqref{eq:procrustes}, it is easy to show that this Procrustes-type problem has a closed form solution in terms of a singular value decomposition. Namely, the solution to \eqref{eq:procrustes_gw_problem} is $\P^* = \U\V^\top$, where $\U\Sigma\V^* = \X_{1:m}\Gamma^*\Y_{1:m}^\top$. After obtaining this projection, we can immediately map the rest of the embeddings via $\hat{\y}^{(j)} = \P^* \y^{(j)}$. 

We point out that this two-step procedure resembles that of \citet{conneau2018word}. Both ultimately produce an orthogonal mapping obtained by solving a Procrustes problems, but they differ in the way they produce pseudo-matches to allow for such second-step: while their approach relies on an adversarially-learned transformation, we use an explicit optimization problem. 

We end this section by discussing parameter and configuration choices. To leverage the fast algorithm of \citet{peyre2016gromov}, we always use the $L_2$ distance as the loss function $L$ between cost matrices. On the other hand, we observed throughout our experiments that the choice of cosine distance as the metric in both spaces consistently leads to better results, which agrees with common wisdom on computing distances between word embeddings. This leaves us with a single hyper-parameter to control: the entropy regularization term $\lambda$. By applying any sensible normalization on the cost matrices (e.g., dividing by the mean or median value), we are able to almost entirely eliminate sensitivity to that parameter. In practice, we use a simple scheme in all experiments: we first try the same fixed value ($\lambda=\num{5e-5}$), and if the regularization proves too small (by leading to floating point errors), we instead use $\lambda=\num{1e-4}$. We never had to go beyond these two values in all our experiments. We emphasize that at no point we use train (let alone test) supervision available with many datasets---model selection is done solely in terms of the unsupervised objective. Pseudocode for the full method (with $L=L_2$ and cosine similarity) is shown here as Algorithm~\ref{alg:GW}. 

% subsection scaling_up (end)

\begin{algorithm}[t]
    \caption{Gromov-Wasserstein Computation for Word Embedding Alignment}\label{alg:GW}
\begin{algorithmic}
   \STATE {\bfseries Input:} Source and target embeddings $\X$, $\Y$. Regularization $\lambda$. Probability vectors $\p,\q$.
   \STATE // Compute intra-language similarities
   \STATE $\C_s \gets \cos(\X,\X), \quad\C_t \gets \cos(\Y,\Y)$
   \STATE $\C_{st} \gets \C_s^2\p\one_{m}^\top + \one_n\q(\C_t^2)^\top$
   \WHILE{not converged}
   		\STATE // Compute pseudo-cost matrix (Eq. \eqref{eq:gw_pseudocost}) 
   		\STATE $\hat{\C}_\Gamma \gets \C_{st} - 2\C_s\Gamma \C_t^\top$
	    \STATE // Sinkhorn iterations (Eq. \eqref{eq:sikhorn_iters})
		\STATE $\a \gets \one, \quad \K \gets \exp\{-\hat{\C}_\Gamma /\lambda\}$
	    \WHILE{not converged}
			\STATE $\a \gets \p \oslash \K \b, \medspace \b \gets \q \oslash \K^\top \a$
		\ENDWHILE
		\STATE $\Gamma \gets \diag{\a}\K\diag{\b}$
   \ENDWHILE
   \STATE // Optional step: Learn explicit projection
	\STATE $\U, \Sigma, \V^\top \gets \text{SVD}(\X\Gamma\Y^\top)$  
	\STATE $\P = \U\V^\top$
	\RETURN $\Gamma, \P$	
\end{algorithmic}
\end{algorithm}

\begin{figure*}[!ht]
	\centering
    \begin{subfigure}[t]{0.33\linewidth}
        \centering
		\includegraphics[width=\linewidth]{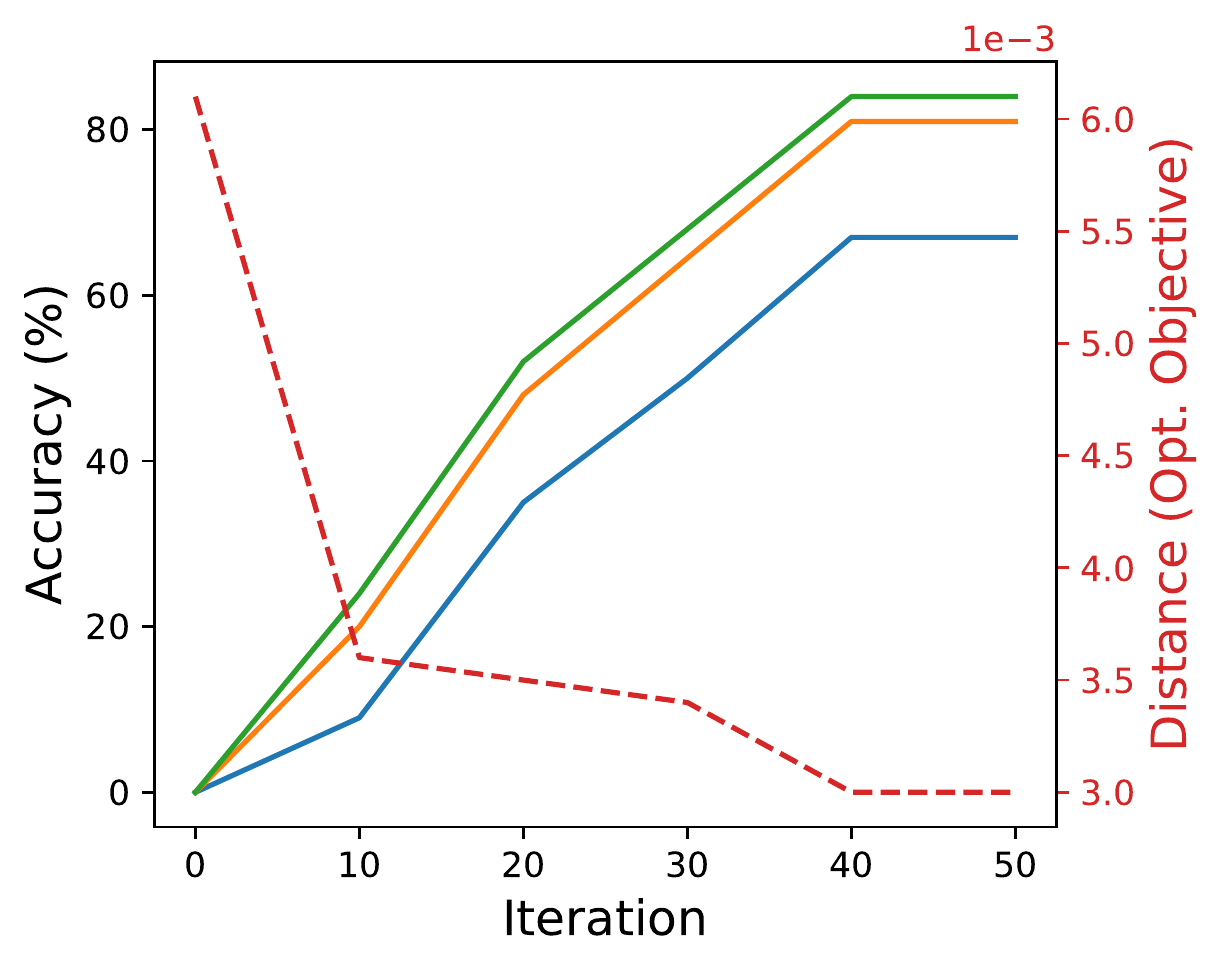}%	
        \caption{\En$\rightarrow$\Fr, 15K words, $\lambda=5\cdot10^{-4}$}\label{fig:dyn_fr_bad}
    \end{subfigure}%
	~	
    \begin{subfigure}[t]{0.33\linewidth}
        \centering
		\includegraphics[width=\linewidth]{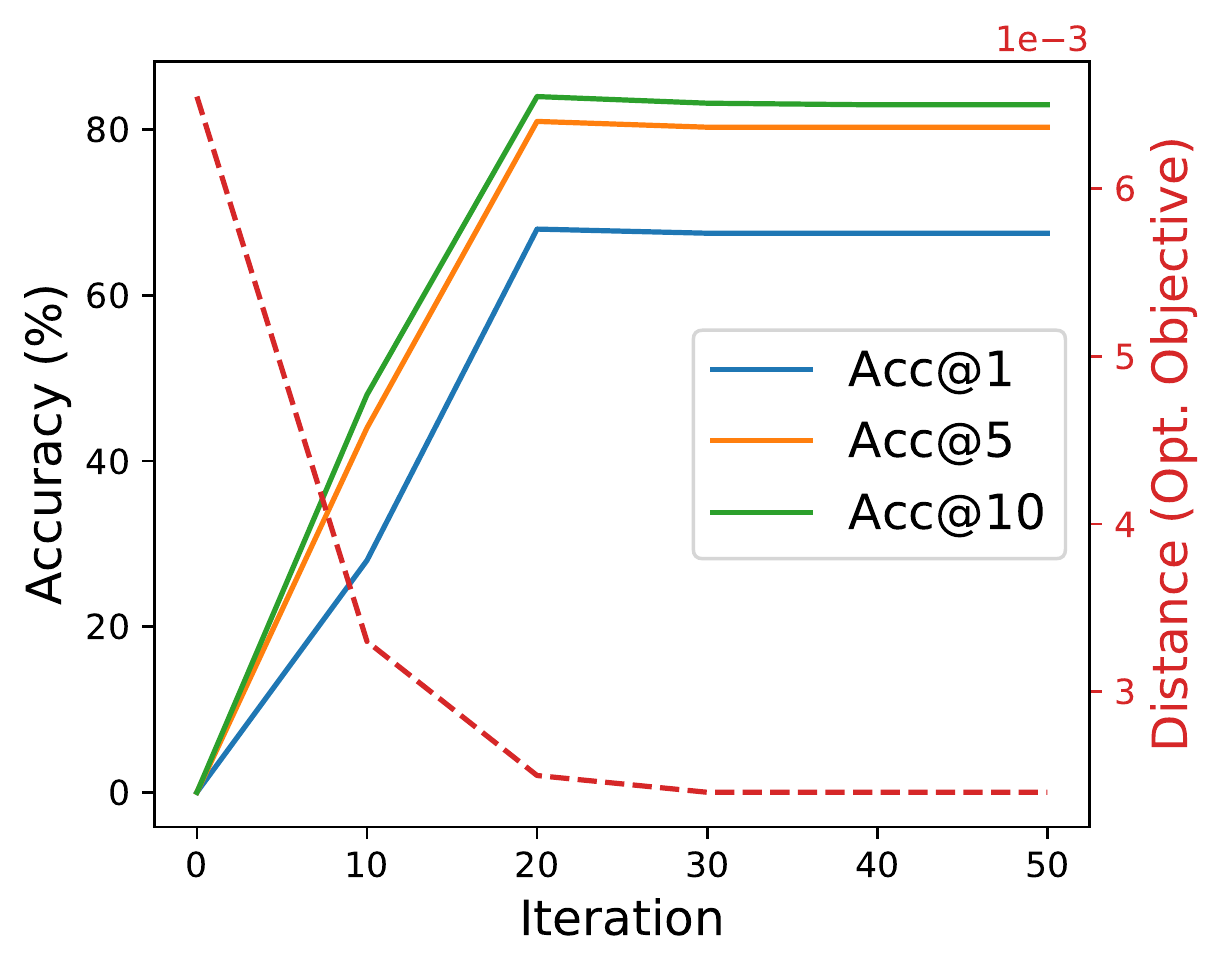}%	
        \caption{\En$\rightarrow$\Fr, 15K words, $\lambda=10^{-4}$}\label{fig:dyn_fr_good}
    \end{subfigure}%
	~
    \begin{subfigure}[t]{0.33\linewidth}
        \centering
		\includegraphics[width=\linewidth]{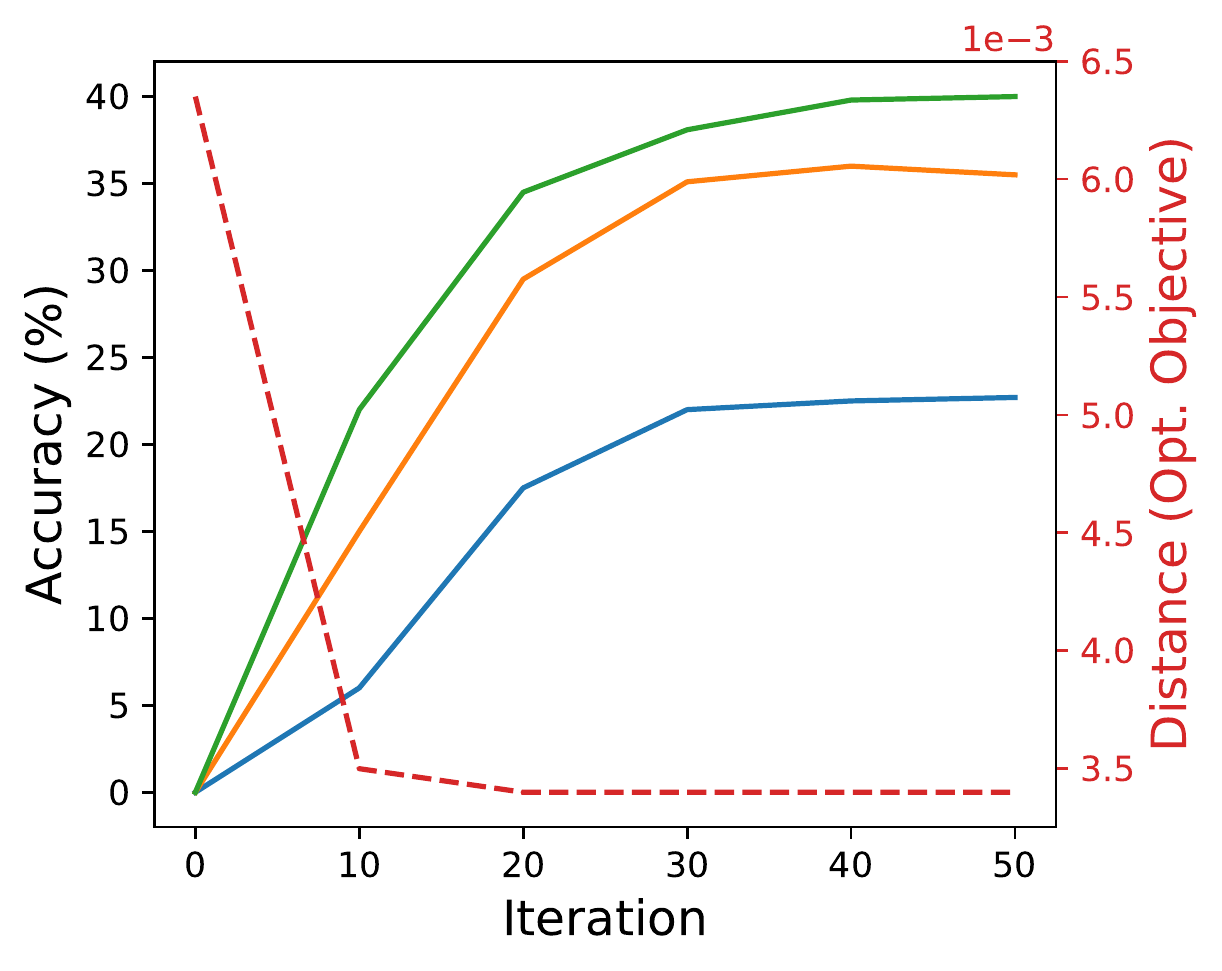}%	
        \caption{\En$\rightarrow$\Ru, 15K words, $\lambda=10^{-4}$}\label{fig:dyn_ru}
    \end{subfigure}%
	\caption{\textbf{Training dynamics for the Gromov-Wasserstein alignment problem}. The algorithm provably makes progress in each iteration, and the objective (red dashed line) closely follows the metric of interest (translation accuracy, not available during training). More related languages (e.g., \En$\rightarrow$\Fr in \ref{fig:dyn_fr_good},\ref{fig:dyn_fr_bad}) lead to faster optimization, while more distant pairs yield slower learning curves (\En$\rightarrow$\Ru, \ref{fig:dyn_ru}). }\label{fig:dynamics}
\end{figure*}

\section{Experiments}\label{sec:experiments} 

Through this experimental evaluation we seek to: (i) understand the optimization dynamics of the proposed approach (§\ref{sub:dynamics}), evaluate its performance on benchmark cross-lingual word embedding tasks (§\ref{sub:benchmark_results}), and (iii) qualitatively investigate the notion of distance-between-languages it computes (§\ref{sub:qualitative_results}). Rather than focusing solely on prediction accuracy, we seek to demonstrate that the proposed approach offers a fast, principled, and robust alternative to state-of-the-art multi-step methods, delivering comparable performance. 

\subsection{Evaluation Tasks and Methods} % (fold)
\label{sub:evaluation_tasks}

\paragraph{Datasets} We evaluate our method on two standard benchmark tasks for cross-lingual embeddings. First, we consider the dataset of \citet{conneau2018word}, which consists of word embeddings trained with \textsc{fastText} \cite{bojanowski2017enriching} on Wikipedia and parallel dictionaries for 110 language pairs. Here, we focus on the language pairs for which they report results: English (\En) from/to Spanish (\Es), French (\Fr), German (\De), Russian (\Ru) and simplified Chinese (\Zh). We do not report results on Esperanto (\Eo) as dictionaries for that language were not provided with the original dataset release.

For our second set of experiments, we consider the---substantially harder\footnote{We discuss the difference in hardness of these two benchmark datasets in Section~\ref{sub:benchmark_results}.}---dataset of \cite{dinu2014improving}, which has been extensively compared against in previous work. It consists of embeddings and dictionaries in four pairs of languages; \En from/to \Es, \It, \De, and \Fi (Finnish).

\paragraph{Methods} To see how our fully-unsupervised method compares with methods that require (some) cross-lingual supervision, we follow \citep{conneau2018word} and consider a simple but strong baseline consisting of solving a procrustes problem directly using the available cross-lingual embedding pairs. We refer to this method simply as \procru. In addition, we compare against the fully-unsupervised methods of \citet{zhang2017adversarial}, \citet{Artetxe2018Robust} and \citet{conneau2018word}.\footnote{Despite its relevance, we do not include the OT-based method of \citet{zhang2017earth} in the comparison because their implementation required use of proprietary software.} As proposed by the latter, we use \textsc{Csls} whenever nearest neighbor search is required, which has been shown to improve upon naive nearest-neighbor retrieval in multiple work.

\begin{table*}[!ht]
	\footnotesize
  \centering
  \begin{tabular}{l l c c c c c c c c c c c c c c}
    \toprule
 	   & & & \multicolumn{2}{c}{\En-\Es} & \multicolumn{2}{c}{\En-\Fr}  & \multicolumn{2}{c}{\En-\De}  & \multicolumn{2}{c}{\En-\It}  & \multicolumn{2}{c}{\En-\Ru} \\
	   \cmidrule(lr){4-5} \cmidrule(lr){6-7}  \cmidrule(lr){8-9}  \cmidrule(lr){10-11}  \cmidrule(lr){12-13}
	   & Supervision & Time & $\rightarrow$ & $\leftarrow$ & $\rightarrow$ & $\leftarrow$ & $\rightarrow$ & $\leftarrow$ & $\rightarrow$ & $\leftarrow$ & $\rightarrow$ & $\leftarrow$  \\
	    \midrule
		\procru  & 5K words & 3 & 77.6 & 77.2 & 74.9 & 75.9 & 68.4 & 67.7 & 73.9 & 73.8 & 47.2 & 58.2\\
		\procru + CSLS & 5K words & 3 & 81.2 & 82.3 & 81.2 & 82.2 & 73.6 & 71.9 & 76.3 & \textbf{75.5} & 51.7 & 63.7 \\
	    \citep{conneau2018word} &  None  & 957 & \textbf{81.7} & \textbf{83.3} & \textbf{82.3} & 82.1 & 74.0 & 72.2 & 77.4 & 76.1& \textbf{52.4} & \textbf{61.4} \\
		% \citep{Artetxe2018Robust} & & 48.53 & 8.9 & 48.47 & 7.3 & 33.50 & 12.9 & 37.60 & 9.1 \\
		\midrule		
		\ours $(\lambda = 10^{-4})$&  None & 70 & 78.3 & 79.5 & 79.3 & 78.3 & 69.6 & 66.9 & 75.3  & 74.1 & 26.1 & 35.4 \\		
		\ours $(\lambda = 10^{-5})$&  None & 37 & \textbf{81.7} & 80.4 &  81.3 & 78.9 & 71.9 & \textbf{72.8} & \textbf{78.9} & 75.2 & 45.1 & 43.7 \\		
    \bottomrule
  \end{tabular}
  \caption{Performance (P@1) of unsupervised and minimally-supervised methods on the dataset of \citet{conneau2018word}.  The time columns shows the average runtime in minutes of an instance (i.e., one language pair) of the method in this task on the same quad-core \textbf{CPU} machine.}
  \label{tab:conneau_results}
\end{table*}

\subsection{Training Dynamics of \ours}
\label{sub:dynamics}
As previously mentioned, our approach involves only two optimization choices, one of which is required only for very large settings. When running Algorithm~\ref{alg:GW} for the full set of embeddings is infeasible (due to memory limitations), one must decide what fraction of the embeddings to use during optimization. In our experiments, we use the largest possible size allowed by memory constraints, which was found to be $K=20,000$ for the personal computer we used. 

The other---more interesting---optimization choice involves the entropy regularization parameter $\lambda$ used within the Sinkhorn iterations. Large regularization values lead to denser optimal coupling $\Gamma^*$, while less regularization leads to sparser solutions,\footnote{In the limit $\lambda\rightarrow 0$, when $n=m$, the solution converges to a permutation matrix, which gives a hard-matching solution to the transportation problem \citep{Peyre2018Computational}.} at the cost of a harder (more non-convex) optimization problem. 

In Figure~\ref{fig:dynamics} we show the training dynamics of our method when learning correspondences between word embeddings from the dataset of \citet{conneau2018word}. As expected, larger values of $\lambda$ lead to smoother improvements with faster runtime-per-iteration, at a price of some drop in performance. In addition, we found that computing GW distances between closer languages (such as \En and \Fr) leads to faster convergence than for more distant ones (such as \En and \Ru, in Fig.~\ref{fig:dyn_ru}).

Worth emphasizing are three desirable optimization properties that set apart the Gromov-Wasserstein distance from other unsupervised alignment approaches, particularly adversarial-training ones: (i) the objective decreases monotonically (ii) its value closely follows the true metric of interest (translation, which naturally is not available during training) and (iii) there is no risk of degradation due to \emph{overtraining}, as is the case for adversarial-based methods trained with stochastic gradient descent \citep{conneau2018word}.

\subsection{Benchmark Results} % (fold)
\label{sub:benchmark_results}

\begin{figure}[t]
	\centering
	\includegraphics[trim={0.3cm 0 13cm 0}, clip, scale=0.3]{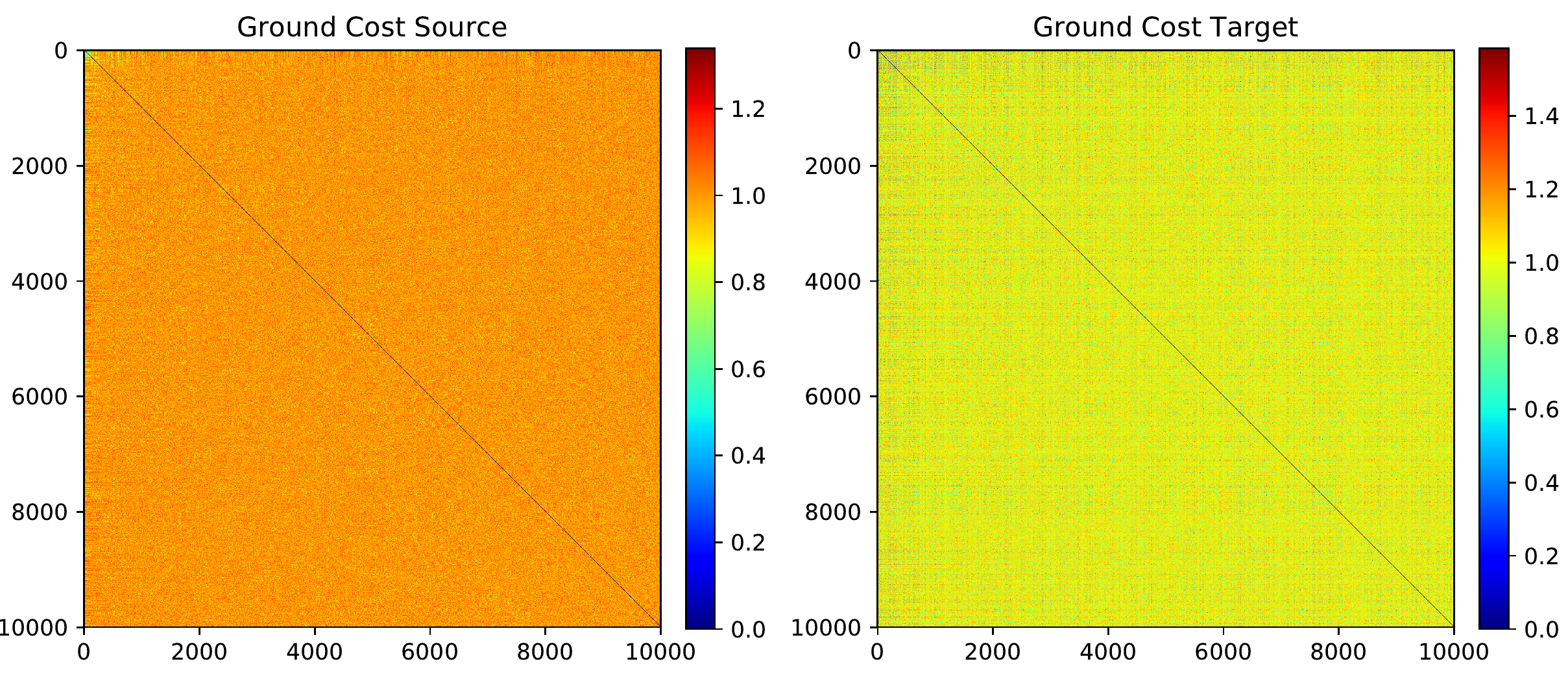}
	\includegraphics[trim={0.3cm 0 13cm 0}, clip, scale=0.3]{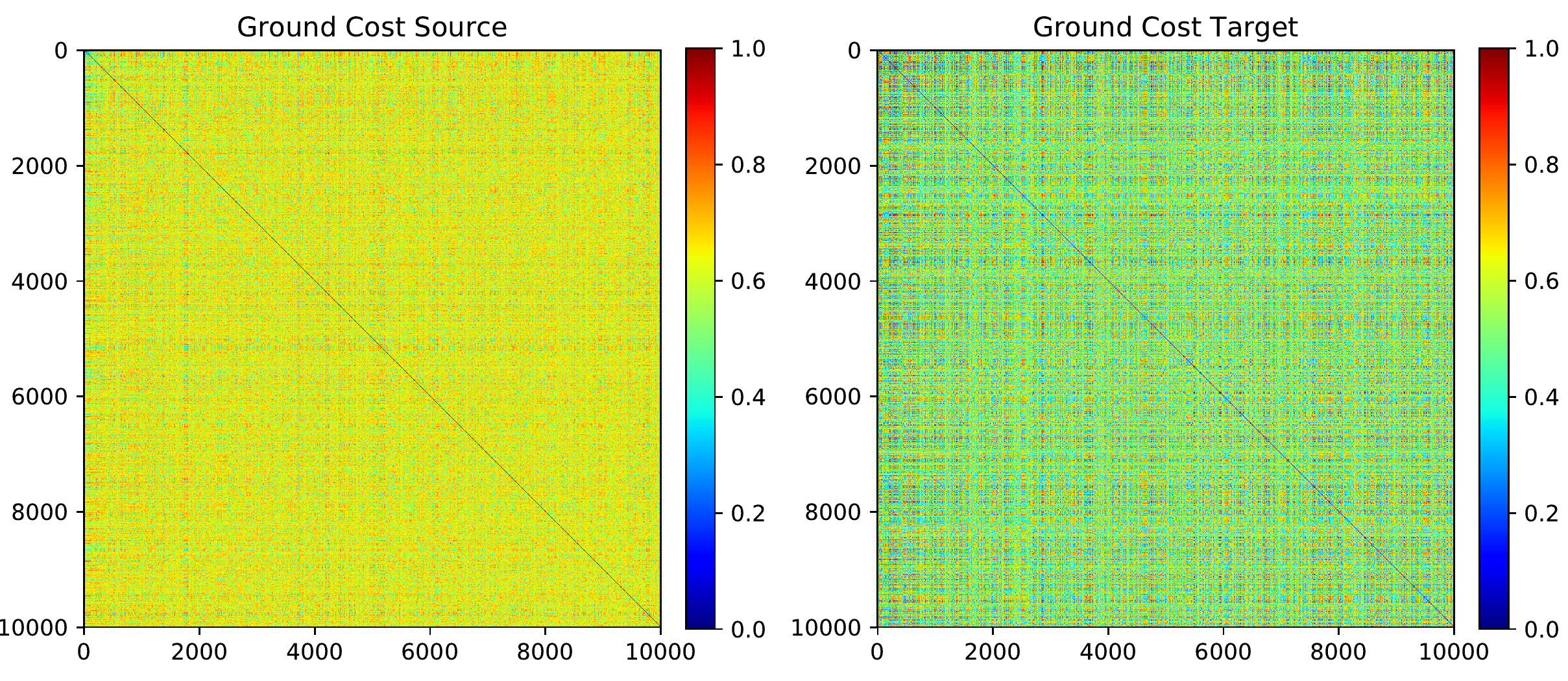}%	
	\caption{\textbf{Top}: Word embeddings trained on non-comparable corpora can lead to uneven distributions of pairwise distances as shown here for the \En-\Fi pair of \cite{dinu2014improving}. \textbf{Bottom:} Normalizing the cost matrices leads to better optimization and improved performance.}\label{fig:cost_assymetry}
\end{figure}

We report the results on the dataset of \citet{conneau2018word} in Table~\ref{tab:conneau_results}. The strikingly high performance of all methods on this task belies the hardness of the general problem of unsupervised cross-lingual alignment. Indeed, as pointed out by \citet{Artetxe2018Robust}, the \textsc{fastText} embeddings provided in this task are trained on very large and highly comparable---across languages---corpora (Wikipedia), and focuses on closely related pairs of languages. Nevertheless, we carry out experiments here to have a broad evaluation of our approach in both \emph{easier} and \emph{harder} settings.

\begin{table*}[t]
  \centering
	%\small
  \begin{tabular}{l c c c c c c c c }
    \toprule
 	   &  \multicolumn{2}{c}{\En-\It} & \multicolumn{2}{c}{\En-\De}  & \multicolumn{2}{c}{\En-\Fi}  & \multicolumn{2}{c}{\En-\Es} \\
	   \cmidrule(lr){2-3} \cmidrule(lr){4-5}  \cmidrule(lr){6-7}  \cmidrule(lr){8-9}
	   & P@1 & Time & P@1 & Time & P@1 & Time & P@1 & Time \\
	    \midrule
		\cite{zhang2017adversarial}$\dagger$ & 0 & 46.6 & 0 & 46.0 & 0.07 & 44.9 & 0.07 & 43.0 \\
	    \citep{conneau2018word}$\dagger$ & 45.40  & 46.1 & 47.27 & 45.4 & 1.62 & 44.4  & 36.20 & 45.3\\
		\citep{Artetxe2018Robust}$\dagger$  & 48.53 & 8.9 & \textbf{48.47} & 7.3 & \textbf{33.50 }& 12.9 & \textbf{37.60} & 9.1 \\
		\midrule
		\ours  & 44.4 & 35.2 & 37.83 & 36.7 & 6.8 & 15.6 & 12.5 & 18.4 \\
		\ours + \textsc{Normalize}  & \textbf{49.21} & 36 & 46.5 & 33.2 & 18.3 & 42.1 & \textbf{37.60} & 38.2\\		
    \bottomrule
  \end{tabular}
  \caption{Results of unsupervised methods on the dataset of \citet{dinu2014improving} with runtimes in minutes. Those marked with $\dagger$ are from \citep{Artetxe2018Robust}. Note that their runtimes correspond to GPU computation, while ours are CPU-minutes, so the numbers are not directly comparable.}\label{tab:dinu_results}
\end{table*}

Next, we present results on the more challenging dataset of \cite{dinu2014improving} in Table~\ref{tab:dinu_results}. Here, we rely on the results reported by \citep{Artetxe2018Robust} since by the time of writing the present work their implementation was not available yet. 

Part of what makes this dataset hard is the wide discrepancy between word distance across languages, which translates into uneven distance matrices (Figure~\ref{fig:cost_assymetry}), and in turn leads to poor results for \ours. To account for this, previous work has relied on an initial whitening step on the embeddings. In our case, it suffices to normalize the pairwise similarity matrices to the same range to obtain substantially better results. While we have observed that careful choice of the regularization parameter $\lambda$ can obviate the need for this step, we opt for the normalization approach since it allows us to optimize without having to tune $\lambda$.
 
We compare our method (with and without normalization) against alternative approaches in Table~\ref{tab:dinu_results}. Note that we report the runtimes of \citet{Artetxe2018Robust} as-is, which are obtained by running on a Titan XP GPU, while our runtimes are, as before, obtained purely by CPU computation.

\subsection{Qualitative Results}\label{sub:extra_exp} % (fold)
\label{sub:qualitative_results}

\begin{figure}[t]
	\centering
	\includegraphics[width=\linewidth]{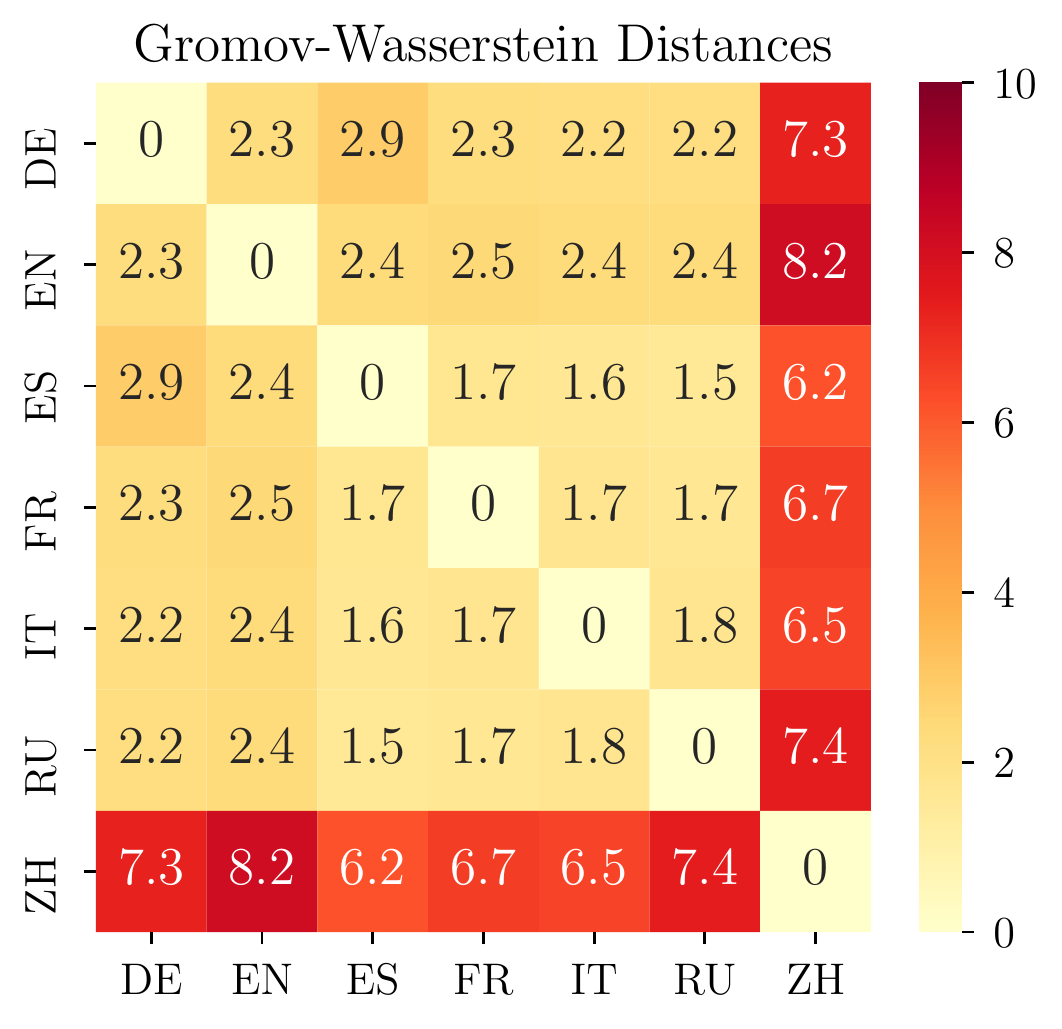}%
	\caption{Pairwise language Gromov-Wasserstein distances obtained as the minimal transportation cost \eqref{eq:gromov_wasserstein} between word embedding similarity matrices. Values scaled by $10^{2}$ for easy visualization.}\label{fig:distance_matrix}
\end{figure}

As mentioned earlier, Theorem~\ref{thm:memoli} implies that the optimal value of the Gromov-Wasserstein problem can be legitimately interpreted as a distance between languages, or more explicitly, between their word embedding spaces. This distributional notion of distance is completely determined by pair-wise geometric relations between these vectors. In Figure~\ref{fig:distance_matrix} we show the values $\gw(\C_s,\C_t, \p,\q)$ computed on the \textsc{fastText} word embeddings of \citet{conneau2018word} corresponding to the most frequent $2000$ words in each language. 

Overall, these distances conform to our intuitions: the cluster of romance languages exhibits some of the shortest distances, while classical Chinese (\Zh) has the overall largest discrepancy with all other languages. But somewhat surprisingly, Russian is relatively close to the romance languages in this metric. We conjecture that this could be due to Russian's rich morphology (a trait shared by romance languages but not English). Furthermore, both Russian and Spanish are pro-drop languages \cite{Haspelmath2001European} and share syntactic phenomena, such as dative subjects \citep{Moore2000What, melis2013historical} and differential object marking \citep{bossong1991differential}, which might explain why \Es is closest to \Ru overall.

On the other hand, English appears remarkably isolated from all languages, equally distant from its germanic (\De) and romance (\Fr) cousins. Indeed, other aspects of the data (such as corpus size) might be underlying these observations.

% subsection qualitative_resuls (end)
% subsection results (end)

%%% Local Variables:
%%% mode: latex
%%% TeX-master: "main"
%%% End:
		% Section 5
%!TEX root = main.tex

\section{Related Work}

Study of the problem of bilingual lexical induction goes back to \citet{rapp1995identifying} and \citet{fung1995compiling}. While the literature on this topic is extensive, we focus here on recent fully-unsupervised and minimally-supervised approaches, and refer the reader to one of various existing surveys for a broader panorama \citep{upadhyay2016cross, ruder2017survey}.

\paragraph{Methods with coarse or limited parallel data.} Most of these fall in one of two categories: methods that learn a mapping from one space to the other, e.g., as a least-squares objective (e.g., \citep{Mikolov2013Exploiting}) or via orthogonal transformations \citet{zhang2016ten, smith2017offline, artetxe2016learning}, and methods that find a common space on which to project both sets of embeddings \cite{faruqui2014improving, lu2015deep}.

\paragraph{Fully Unsupervised methods.} % (fold)
\label{par:fully_unsupervised_methods}

\citet{conneau2018word} and \citet{zhang2017adversarial} rely on adversarial training to produce an initial alignment between the spaces. The former use pseudo-matches derived from this initial alignment to solve a Procrustes \eqref{eq:procrustes} alignment problem. Our Gromov-Wasserstein framework can be thought of as providing an alternative to these adversarial training steps, albeit with a concise optimization formulation and producing explicit matches (via the optimal coupling) instead of depending on nearest neighbor search, as the adversarially-learnt mappings do. 

\citet{zhang2017earth} also leverage optimal transport distances for the cross-lingual embedding task. However, to address the issue of non-alignment of embedding spaces, their approach follows the joint optimization of the transportation and procrustes problem as outlined in Section~\ref{sub:ot}.  This formulation makes an explicit modeling assumption (invariance to unitary transformations), and requires repeated solution of Procrustes problems during alternating minimization. Gromov-Wasserstein, on the other hand, is more flexible and makes no such assumption, since it directly deals with similarities rather than vectors. In the case where it is required, such an orthogonal mapping can be obtained by solving a single procrustes problem, as discussed in Section~\ref{sub:scaling_up}.

\section{Discussion and future work}
In this work we provided a direct optimization approach to cross-lingual word alignment. The Gromov-Wasserstein distance is well-suited for this task as it performs a relational comparison of word-vectors across languages rather than word-vectors directly. The resulting objective is concise, and can be optimized efficiently. The experimental results show that the resulting alignment framework is fast, stable and robust, yielding near state-of-the-art performance at a computational cost orders of magnitude lower than that of alternative fully unsupervised methods. 

While directly solving Gromov-Wasserstein problems of reasonable size is feasible, scaling up to large vocabularies made it necessary to learn an explicit mapping via Procrustes. GPU computations or stochastic optimization could help avoid this secondary step.

\section*{Acknowledgments}

The authors would like to thank the anonymous reviewers for helpful feedback. The work was partially supported by MIT-IBM grant ``Adversarial learning of multimodal and structured data'', and Graduate Fellowships from Hewlett Packard and CONACYT.

%%% Local Variables:
%%% mode: latex
%%% TeX-master: "main"
%%% End:
		% Related + Conclusion

\bibliographystyle{acl_natbib_nourl}
\bibliography{OrthogonalOT.bib}

\begin{thebibliography}{31}
\expandafter\ifx\csname natexlab\endcsname\relax\def\natexlab#1{#1}\fi

\bibitem[{Artetxe et~al.(2016)Artetxe, Labaka, and
  Agirre}]{artetxe2016learning}
Mikel Artetxe, Gorka Labaka, and Eneko Agirre. 2016.
\newblock {Learning principled bilingual mappings of word embeddings while
  preserving monolingual invariance}.
\newblock In \emph{Proceedings of the 2016 Conference on Empirical Methods in
  Natural Language Processing}, pages 2289--2294.

\bibitem[{Artetxe et~al.(2018)Artetxe, Labaka, and Agirre}]{Artetxe2018Robust}
Mikel Artetxe, Gorka Labaka, and Eneko Agirre. 2018.
\newblock {A robust self-learning method for fully unsupervised cross-lingual
  mappings of word embeddings}.
\newblock In \emph{Proceedings of the 56th Annual Meeting of the Association
  for Computational Linguistics (Volume 1: Long Papers)}, pages 789----798.
  Association for Computational Linguistics.

\bibitem[{Bojanowski et~al.(2017)Bojanowski, Grave, Joulin, and
  Mikolov}]{bojanowski2017enriching}
Piotr Bojanowski, Edouard Grave, Armand Joulin, and Tomas Mikolov. 2017.
\newblock {Enriching Word Vectors with Subword Information}.
\newblock \emph{Transactions of the Association for Computational Linguistics},
  5:135--146.

\bibitem[{Bossong(1991)}]{bossong1991differential}
Georg Bossong. 1991.
\newblock {Differential object marking in Romance and beyond}.
\newblock \emph{New analyses in Romance linguistics}, pages 143--170.

\bibitem[{Conneau et~al.(2018)Conneau, Lample, Ranzato, Denoyer, and
  J{\'{e}}gou}]{conneau2018word}
Alexis Conneau, Guillaume Lample, Marc'Aurelio Ranzato, Ludovic Denoyer, and
  Herv{\'{e}} J{\'{e}}gou. 2018.
\newblock {Word Translation Without Parallel Data}.
\newblock In \emph{International Conference on Learning Representations}.

\bibitem[{Cuturi(2013)}]{cuturi2013sinkhorn}
Marco Cuturi. 2013.
\newblock {Sinkhorn distances: Lightspeed computation of optimal transport}.
\newblock In \emph{Advances in Neural Information Processing Systems}, pages
  2292----2300.

\bibitem[{Dinu et~al.(2014)Dinu, Lazaridou, and Baroni}]{dinu2014improving}
Georgiana Dinu, Angeliki Lazaridou, and Marco Baroni. 2014.
\newblock {Improving zero-shot learning by mitigating the hubness problem}.
\newblock \emph{arXiv preprint arXiv:1412.6568}.

\bibitem[{Faruqui and Dyer(2014)}]{faruqui2014improving}
Manaal Faruqui and Chris Dyer. 2014.
\newblock {Improving vector space word representations using multilingual
  correlation}.
\newblock In \emph{Proceedings of the 14th Conference of the European Chapter
  of the Association for Computational Linguistics}, pages 462--471.

\bibitem[{Fung(1995)}]{fung1995compiling}
Pascale Fung. 1995.
\newblock {Compiling bilingual lexicon entries from a non-parallel
  English-Chinese corpus}.
\newblock In \emph{Third Workshop on Very Large Corpora}.

\bibitem[{Guo et~al.(2015)Guo, Che, Yarowsky, Wang, and Liu}]{guo2015cross}
Jiang Guo, Wanxiang Che, David Yarowsky, Haifeng Wang, and Ting Liu. 2015.
\newblock {Cross-lingual dependency parsing based on distributed
  representations}.
\newblock In \emph{Proceedings of the 53rd Annual Meeting of the Association
  for Computational Linguistics and the 7th International Joint Conference on
  Natural Language Processing (Volume 1: Long Papers)}, volume~1, pages
  1234--1244.

\bibitem[{Hashimoto et~al.(2016)Hashimoto, Alvarez-Melis, and
  Jaakkola}]{Hashimoto2016Word}
Tatsunori~B Hashimoto, David Alvarez-Melis, and Tommi~S Jaakkola. 2016.
\newblock {Word Embeddings as Metric Recovery in Semantic Spaces}.
\newblock \emph{Transactions of the Association for Computational Linguistics},
  4:273--286.

\bibitem[{Haspelmath(2001)}]{Haspelmath2001European}
Martin Haspelmath. 2001.
\newblock {The European linguistic area: Standard Average European}.
\newblock In \emph{Language typology and language universals: An international
  handbook}, volume~2, pages 1492--1510. de Gruyter.

\bibitem[{Lample et~al.(2018)Lample, Denoyer, and
  Ranzato}]{lample2018unsupervised}
Guillaume Lample, Ludovic Denoyer, and Marc'Aurelio Ranzato. 2018.
\newblock {Unsupervised Machine Translation Using Monolingual Corpora Only}.
\newblock \emph{International Conference on Learning Representations}.

\bibitem[{Lu et~al.(2015)Lu, Wang, Bansal, Gimpel, and Livescu}]{lu2015deep}
Ang Lu, Weiran Wang, Mohit Bansal, Kevin Gimpel, and Karen Livescu. 2015.
\newblock {Deep multilingual correlation for improved word embeddings}.
\newblock In \emph{Proceedings of the 2015 Conference of the North American
  Chapter of the Association for Computational Linguistics: Human Language
  Technologies}, pages 250--256.

\bibitem[{Melis et~al.(2013)Melis, Flores, and Holvoet}]{melis2013historical}
Chantal Melis, Marcela Flores, and A~Holvoet. 2013.
\newblock {On the historical expansion of non-canonically marked ‘subjects'
  in Spanish}.
\newblock \emph{The diachronic Typology of Non-Canonical Subjects,
  Amsterdam/Philadelphia, Benjamins}, pages 163--184.

\bibitem[{M{\'{e}}moli(2011)}]{memoli2011gromov}
Facundo M{\'{e}}moli. 2011.
\newblock {Gromov--Wasserstein distances and the metric approach to object
  matching}.
\newblock \emph{Foundations of computational mathematics}, 11(4):417--487.

\bibitem[{Mikolov et~al.(2013)Mikolov, Le, and
  Sutskever}]{Mikolov2013Exploiting}
Tomas Mikolov, Quoc~V Le, and Ilya Sutskever. 2013.
\newblock {Exploiting Similarities among Languages for Machine Translation}.
\newblock \emph{arXiv preprint arXiv:1309.4168v1}, pages 1--10.

\bibitem[{Moore and Perlmutter(2000)}]{Moore2000What}
John Moore and David~M. Perlmutter. 2000.
\newblock {What does it take to be a dative subject?}
\newblock \emph{Natural Language and Linguistic Theory}, 18(2):373--416.

\bibitem[{Peyr{\'{e}} and Cuturi(2018)}]{Peyre2018Computational}
Gabriel Peyr{\'{e}} and Marco Cuturi. 2018.
\newblock {Computational Optimal Transport}.
\newblock Technical report.

\bibitem[{Peyr{\'{e}} et~al.(2016)Peyr{\'{e}}, Cuturi, and
  Solomon}]{peyre2016gromov}
Gabriel Peyr{\'{e}}, Marco Cuturi, and Justin Solomon. 2016.
\newblock {Gromov-Wasserstein averaging of kernel and distance matrices}.
\newblock In \emph{International Conference on Machine Learning}, pages
  2664--2672.

\bibitem[{Piantadosi(2014)}]{Piantadosi2014Zipf}
Steven~T Piantadosi. 2014.
\newblock {Zipf's word frequency law in natural language: A critical review and
  future directions}.
\newblock \emph{Psychonomic Bulletin {\&} Review}, 21(5):1112--1130.

\bibitem[{Rangarajan et~al.(1997)Rangarajan, Chui, and
  Bookstein}]{Rangarajan1997Softassign}
Anand Rangarajan, Haili Chui, and Fred~L Bookstein. 1997.
\newblock {The Softassign Procrustes Matching Algorithm}.
\newblock \emph{Lecture Notes in Computer Science}, 1230:29--42.

\bibitem[{Rapp(1995)}]{rapp1995identifying}
Reinhard Rapp. 1995.
\newblock {Identifying word translations in non-parallel texts}.
\newblock In \emph{Proceedings of the 33rd annual meeting on Association for
  Computational Linguistics}, pages 320--322. Association for Computational
  Linguistics.

\bibitem[{Rapp(1999)}]{rapp1999automatic}
Reinhard Rapp. 1999.
\newblock {Automatic identification of word translations from unrelated English
  and German corpora}.
\newblock In \emph{Proceedings of the 37th annual meeting of the Association
  for Computational Linguistics on Computational Linguistics}, pages 519--526.
  Association for Computational Linguistics.

\bibitem[{Ruder et~al.(2017)Ruder, Vuli{\'{c}}, and
  S{\o}gaard}]{ruder2017survey}
Sebastian Ruder, Ivan Vuli{\'{c}}, and Anders S{\o}gaard. 2017.
\newblock {A survey of cross-lingual embedding models}.
\newblock \emph{arXiv preprint arXiv:1706.04902}.

\bibitem[{Sch{\"{o}}nemann(1966)}]{schonemann1966generalized}
Peter~H. Sch{\"{o}}nemann. 1966.
\newblock {A generalized solution of the orthogonal procrustes problem}.
\newblock \emph{Psychometrika}, 31(1):1--10.

\bibitem[{Smith et~al.(2017)Smith, Turban, Hamblin, and
  Hammerla}]{smith2017offline}
Samuel~L Smith, David H~P Turban, Steven Hamblin, and Nils~Y Hammerla. 2017.
\newblock {Offline bilingual word vectors, orthogonal transformations and the
  inverted softmax}.
\newblock \emph{International Conference on Learning Representations}.

\bibitem[{Upadhyay et~al.(2016)Upadhyay, Faruqui, Dyer, and
  Roth}]{upadhyay2016cross}
Shyam Upadhyay, Manaal Faruqui, Chris Dyer, and Dan Roth. 2016.
\newblock {Cross-lingual models of word embeddings: An empirical comparison}.
\newblock \emph{arXiv preprint arXiv:1604.00425}.

\bibitem[{Zhang et~al.(2017{\natexlab{a}})Zhang, Liu, Luan, and
  Sun}]{zhang2017adversarial}
Meng Zhang, Yang Liu, Huanbo Luan, and Maosong Sun. 2017{\natexlab{a}}.
\newblock {Adversarial training for unsupervised bilingual lexicon induction}.
\newblock In \emph{Proceedings of the 55th Annual Meeting of the Association
  for Computational Linguistics (Volume 1: Long Papers)}, volume~1, pages
  1959--1970.

\bibitem[{Zhang et~al.(2017{\natexlab{b}})Zhang, Liu, Luan, and
  Sun}]{zhang2017earth}
Meng Zhang, Yang Liu, Huanbo Luan, and Maosong Sun. 2017{\natexlab{b}}.
\newblock {Earth Mover's Distance Minimization for Unsupervised Bilingual
  Lexicon Induction}.
\newblock In \emph{Proceedings of the 2017 Conference on Empirical Methods in
  Natural Language Processing}, pages 1934--1945.

\bibitem[{Zhang et~al.(2016)Zhang, Gaddy, Barzilay, and
  Jaakkola}]{zhang2016ten}
Yuan Zhang, David Gaddy, Regina Barzilay, and Tommi Jaakkola. 2016.
\newblock {Ten Pairs to Tag -- Multilingual POS Tagging via Coarse Mapping
  between Embeddings}.
\newblock In \emph{Proceedings of the 2016 Conference of the North American
  Chapter of the Association for Computational Linguistics: Human Language
  Technologies}, pages 1307--1317, San Diego, California. Association for
  Computational Linguistics.

\end{thebibliography}
%\bibliography{/Users/david/.mendeley_bibfiles/OrthogonalOT.bib}

\end{document}